\documentclass[english]{article}
\usepackage{eucal}
\usepackage[utf8]{inputenc}
\usepackage[T1]{fontenc}
\usepackage{babel}
\usepackage{amsmath}
\usepackage{wrapfig}
\usepackage[svgnames]{xcolor}
\usepackage{float}
\usepackage{amsfonts}
\usepackage{csquotes}
\usepackage{amstext}
\usepackage{amssymb}
\usepackage{graphicx}
\usepackage{comment}
\usepackage{subcaption}
\usepackage{fancyhdr}
\usepackage{siunitx}
\usepackage{hyperref}
\usepackage{authblk}
\hypersetup{
    colorlinks=true,
    linkcolor=blue,
    filecolor=magenta,      
    urlcolor=cyan,
}
\usepackage{graphicx}
\usepackage[para]{footmisc}
\usepackage[backend=biber,style=numeric,sorting=none]{biblatex}

\title{Deep neuroevolution to predict primary brain tumor grade from functional MRI adjacency matrices}
\author[1]{Joseph Stember\thanks{Corresponding Author : joestember@gmail.com}}
\author[1]{Mehrnaz Jenabi}
\author[1]{Luca Pasquini}
\author[1]{Kyung Peck}
\author[1]{Andrei Holodny}
\author[2]{Hrithwik Shalu}

\affil[1]{Department of Radiology, Memorial Sloan Kettering Cancer Center, NY, NY, 10065}
\affil[2]{Department of Aerospace Engineering, Indian Institute of Technology Madras, Chennai, India, 600036}

\date{}

\begin{document}
\maketitle

\thispagestyle{empty}

\begin{abstract}

Whereas MRI produces anatomic information about the brain, functional MRI (fMRI) tells us about neural activity within the brain, including how various regions communicate with each other. The full chorus of conversations within the brain is summarized elegantly in the adjacency matrix. Although information-rich, adjacency matrices typically provide little in the way of intuition. Whereas trained radiologists viewing anatomic MRI can readily distinguish between different kinds of brain cancer, a similar determination using adjacency matrices would exceed any expert's grasp. Artificial intelligence (AI) in radiology usually analyzes anatomic imaging, providing assistance to radiologists. For non-intuitive data types such as adjacency matrices, AI moves beyond the role of helpful assistant, emerging as indispensible. We sought here to show that AI can learn to discern between two important brain tumor types, high-grade glioma (HGG) and low-grade glioma (LGG), based on adjacency matrices. We trained a convolutional neural networks (CNN) with the method of deep neuroevolution (DNE), because of the latter's recent promising results; DNE has produced remarkably accurate CNNs even when relying on small and noisy training sets, or performing nuanced tasks. After training on just 30 adjacency matrices, our CNN could tell HGG apart from LGG with perfect testing set accuracy. Saliency maps revealed that the network learned highly sophisticated and complex features to achieve its success. Hence, we have shown that it is possible for AI to recognize brain tumor type from functional connectivity. In future work, we will apply DNE to other noisy and somewhat cryptic forms of medical data, including further explorations with fMRI. 

\end{abstract}

\pagebreak

\section*{Introduction}

Resting-state functional MRI (rfMRI) provides information about neuronal co-activation when no particular task is required. It is thus not confined to a particular region of the brain, such as pre- and post-central gyrus for motor and sensory tasks such as finger tapping. Also, because of its more holistic nature, rfMRI can provide global information about brain functional connectivity to complement standard anatomic imaging \cite{wang2010graph}. 

A useful goal for rfMRI is classifying between normal and disease states or among different disease states based on functional connectivity \cite{jia2022independent,cao2019toward}. Most research in rfMRI-based classification has focused on classic graph-theoretic metrics such as connectivity \cite{richiardi2011decoding}. A somewhat successful application is distinguishing  autism spectrum disorder (ASD) from neurotypical controls (NC). Because ASD is probably due to the different connectivity of nodes compared to NC, and this is almost a defining characteristic, that metric is both intuitive and lends itself fairly well to threshold-based classification \cite{kessler2016brain,hull2017resting}.

Recent work has automated the classification between ASD and NC using a type of convolutional neural network (CNN) designed for analyzing graphs, the graph neural network (GNN). Deep learning with GNNs was able to reproduce the expected predictions with with high accuracy \cite{li2020pooling,li2021braingnn,wang2021graph,stember2022deep_fmri}. However, despite the success, GNNs have four features, in particular, that may limit their generalizability:
\begin{enumerate}
    \item Convolutional kernels have a maximum  dimension equal to the row or column length in the adjacency matrices or incidence matrices. The intuition is that a given row or column reflects the connection weights between the node represented by the row or column index and all of the other nodes in the network. Hence digesting all of the connected neighbors in a convolutional step would ensure that important connections are not thrown out.
    \item Via the pooling operation, networks can be pruned and simplified in a manner that does not discard intrinsically important information needed for the task of accurate classification. However, the risk remains that useful information could get lost.
    \item GNNs are sensitive to noise and approximations entailed by initial processing, especially if there is spurious signal from a non-involved node. One example is motion artifact, which can falsely suggest communication among neurons. Because rfMRI suffers from low signal-to-noise ratios, such varying stochastic contributions can degrade GNN performance. 
    \item The one-dimensional kernals of GNNs provide less flexibility than two-dimensional kernels. Flexibility with regard to network weights is crucial to learning more tasks, from more data distributions, and with greater generalizability that improves deployment. Flexibility is particularly helpful for evolutionary strategies-based weight tuning, which is the approach we have pursued in the work presented here.
\end{enumerate}

\begin{figure}[H]
\includegraphics[width=12cm]{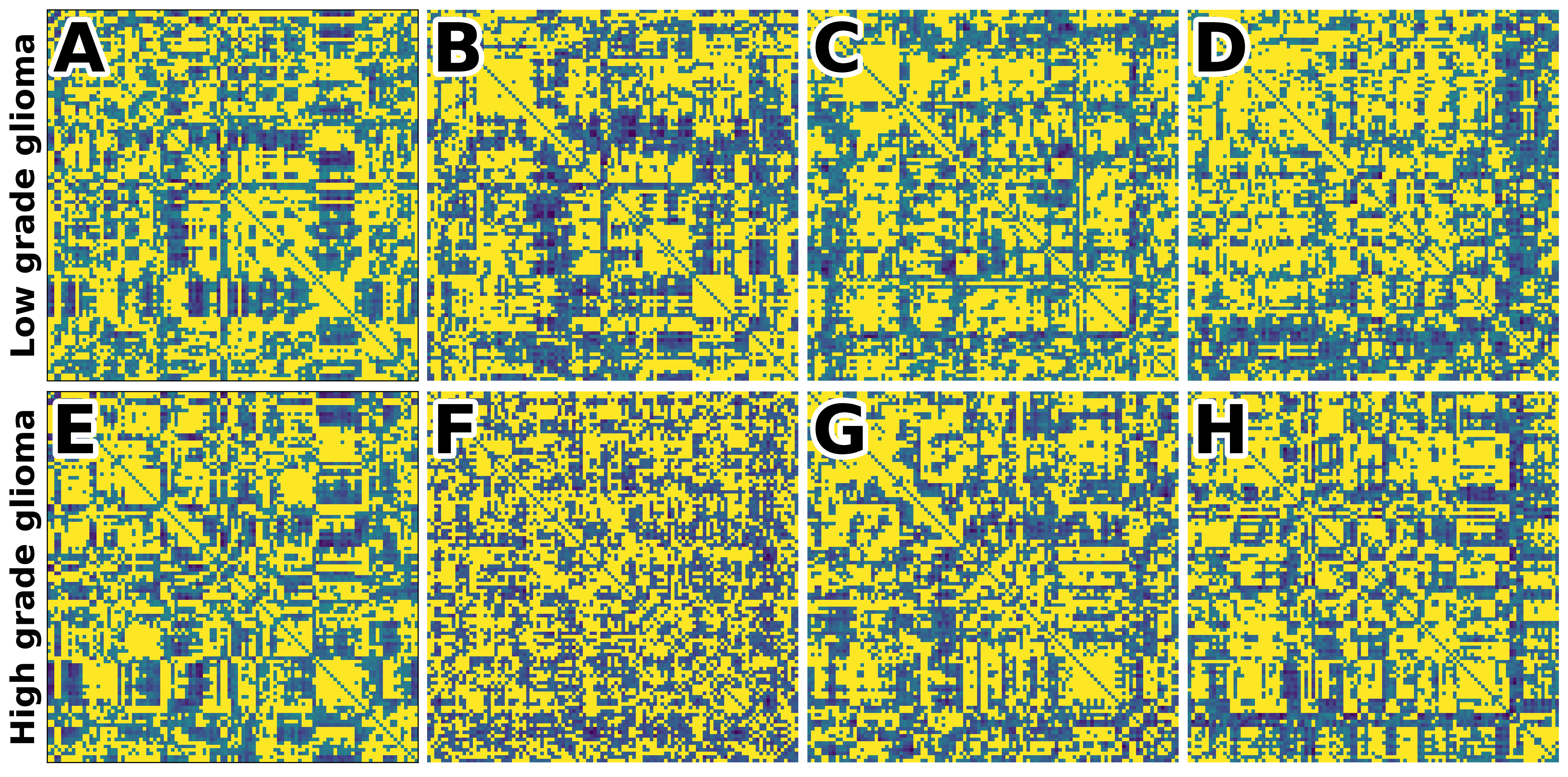}
\caption{Examples of adjacency matrices for low-grade gliomas (LGG) and high-grade gliomas (HGG). All values are scaled from zero to one, the latter represented by yellow pixels.}
\label{fig:adj_mtx_egs}
\end{figure}

The limitations of classic graph theory metrics and GNN-based deep learning present a challenge for tasks that lack the intuition of classifying ASD, in particular, the current work concerns one such task: discerning low-grade gliomas (LGG) from high-grade gliomas (HGG). In  cases such as these, simple connectivity provides no clues that humans can grasp to distinguish the two tumor types reliably. Certainly the high-dimensional hyperplane separating classes is not readily deducible or interpretable by human researchers. Such is the situation for our study, as shown in Figure \ref{fig:adj_mtx_egs}, which displays adjacency matrices derived from rfMRI for a few examples of HGG and LGG. For this data type and task, although differences in connectivity do exist \cite{pasquini2022brain}, no obvious pattern of features that could reliably separate the two classes is apparent from visual inspection.  

Hence, what is needed where differences in functional connectivity are not straightforward is an approach producing adaptive higher-order features, which CNNs are well-suited to learn. In other words, the task requires deep learning. Further, the AI approach must be able to learn from heterogeneous, noisy and complex data that does not conform well to human intuition. 

We have seen that an evolutionary strategies-based approach to train CNNs, called deep neuroevolution (DNE), can learn effectively in the presence of sparse, noisy and heterogeneous data sets \cite{neuroblastoma_DNE}. DNE thus provides a good match for our objectives, because 
\begin{itemize}
    \item DNE is data-efficient, and resistant to overfitting small training sets. fMRI is generally in limited supply, even at academic medical centers.
    \item fMRI adjacency matrices constitute a noisy, non-intuitive form of data.
\end{itemize}
We sought here to apply DNE to discern between adjacency matrices representing high-grade gliomas (HGGs) versus low-grade gliomas (LGGs), recently re-defined by the genomic criteria of isocitrate dehydrogenase mutated (IDH+) versus wild type (IDH-), respectively. We note that IDH+ in most cases is present for histopathologic LGGs, whereas HGGs typically feature IDH- genotypes. We, therefore, take the liberty of continuing to use the titles of LGG and HGG, the former representing IDH+ tumors and the latter being IDH-. 

Having defined our task, we sought to perform DNE-based analysis using small square convolutional kernels, to reflect the widely used and standard method for anatomic images. We also wished to eschew GNN's pooling-based production of simplified coarse-grained graphs as intermediate stages of forward passing, again for simplification and to more fully align our approach with the prevailing style of forward passing that is broadly employed in computer vision CNNs. 

We will describe in this work how we analyzed adjacency matrices that represent functional connectivity between brain regions. In particular, we will discuss being able to train a CNN whose architecture is just like those used analyze anatomic images. We will show that the trained network achieved highly accurate distinction between LGG and HGG. We will briefly introduce the rationale behind DNE and outline its implementation and workflow, noting that DNE itself is further described in other literature. Here we emphasize DNE's ability to learn complex higher-order features that provide reliable predictions of lesion type.
 
\section{Methods}

\subsection{Patients}

The present retrospective cross-sectional study was approved by the institutional review board and conducted in agreement with the Helsinki declaration. Informed consent was waived by the Institutional Review Board, as the study was retrospective. Patients were selected from the archive of our institution from January 2016 to September 2021 according to the following inclusion criteria: newly diagnosed left-hemispheric glioma (according to the World Health Organization 2016 classification \cite{louis20162016}); no prior surgery or other treatment; availability of resting-state data acquired with the same protocol and therefore comparable; absence of tumor-related or patient-related artifacts including drop-out from hemorrhagic tumor components and motion. The average LGG patient age was 40 years, while for HGG average age was 62 years.

\subsection{MRI Acquisition}

rfMRI images were obtained using a 3T scanner (GE Healthcare, Chicago, IL) with 24-channel head coils. fMRI was acquired with single-shot EPI ($TR/TE=2500/33$ msec, slice thickness=4 mm, matrix=64×64, field of view=240 mm, scanning duration: 6 min 55 sec). rfMRI matching FLAIR (TR/TE=10,000/106 ms, inversion time=220 ms, matrix=256×256), T1 post-contrast ($TR/TE=600/20$ ms, 256×256 matrix) images and 3D T1-weighted anatomical images using a spoiled gradient recalled sequence (TR/TE=22/4 ms, matrix=256×256, slice thickness=1 mm) were obtained as routine clinical scans. During the resting-state fMRI scan, subjects were instructed to relax, fixate on a central cross, and try to clear their minds.

Images were pre-processed and analyzed with CONN toolbox \cite{whitfield2012conn} (implemented in SPM (SPM 12, The Wellcome Centre for Human Neuroimaging, UCL Queen Square Institute of Neurology, London, UK) and MATLAB (R2016b)).

For pre-processing: depicting, motion correction, slice timing correction, volume registration, smoothing with a Gaussian filter with a full-width-half-maximum (FWHM) of 6 mm, and linear detrending were applied to raw fMRI data to correct head motion, to suppress noise and to increase the SNR ratio. In addition, rsfMRI data were filtered (0.01–0.08 Hz) to extract functional integration processes and to remove respiratory and cardiac noise during rest.

For analyzing direct normalization of functional and structural data to MNI space plus segmentation of all data to gray matter, white matter and cerebrospinal fluid were applied. The co-registered anatomical and functional images in MNI space were parcellated into 136 region of interest (ROI) (136 parcels) using the default atlas of CONN toolbox \cite{whitfield2012conn}. All coregistered anatomical and functional images were inspected for transformation of tumor area to MNI space to ensure correct anatomical parcellation via the atlas. 

Finally, the average timeseries within each ROIs (parcellated regions, to which we refer interchangeably as nodes) were estimated. ROI-to-ROI connectivity matrices, which we will preferentially refer to as adjacency matrices, represent the connectivity between one of the 136 ROIs to the rest of the ROIs. 

These matrices show the degree of functional connectivity between each pair of ROIs, with each element defined as the Fisher-transformed bivariate correlation coefficient between a pair of ROIs BOLD timeseries \cite{whitfield2012conn}. ROI-to-ROI connectivity matrices were saved for further analyses. 

Because we were only studying supratentorial/cerebral lesions, we excluded the last 30 ROIs, which correspond to the brainstem and cerebellum. This left us with connectivity represented as $105 x 105$ adjacency matrices for each patient. Finally, in order to emphasize strong connections, we set entries in the adjacency matrix that exceeded the 50th percentile equal to one.

We assembled 30 such adjacency matrices for LGG and 30 for HGG. We partitioned this set of adjacency matrices into a training set of 15 LGG and 15 HGG, leaving the other 15 LGG and 15 HGG cases for the testing set. Figure \ref{fig:adj_mtx_egs} shows a few example adjacency matrices. We note here that the means and standard deviations between LGG and HGG were essentially the same. Mean pixel intensity was 0.46 for LGG and 0.45 for HGG, and the standard deviation was 0.29 for LGG and 0.24 for HGG. Hence, these values would not serve to separate LGG from HGG via thresholding.
 
\subsection*{Deep neuroevolution (DNE)}

We sought to train a convolutional neural network (CNN) via an Evolutionary Strategies approach called Deep Neuroevolution (DNE), as opposed to optimizing network weights by the prevailing stochastic gradient descent (SGD) method. DNE has shown remarkable accuracy and generalizability for anatomic images, notably for very small \cite{stember2021_DNE_sequence} and "noisy" \cite{neuroblastoma_DNE} training sets, as well as for use cases entailing complex tasks \cite{stember2022direct}. Because rfMRI adjacency matrices represent a non-intuitive form of data, often difficult for humans to interpret, and in particular because of the complexity of distinguishing between LGG and HGG, we opted to use DNE in the present study. 

The training scheme, described in prior works by Stember and Shalu \cite{stember2021_DNE_sequence,stember2022direct,neuroblastoma_DNE}, replaces SGD's gradient-based weight optimization with random CNN weight permutations, analogous to mutations in an organism's genome. DNE employs a scheme whereby parent CNNs incorporate the best child CNN weights as defined by the selection criterion. In prior work as well as the current study, that selection criterion is the training set classification accuracy. The order of operations provides the key that enables DNE's chief attributes:
\begin{itemize}
    \item Data-efficiency, allowing DNE to learn in a generalized manner on small data sets, so long as the range of training set data reflects that in the testing set or clinical deployment cases.
    \item Noise-resilience, permitting DNE to learn effectively on heterogeneous data from multiple sources with high variation in image characteristics. For example, DNE performed well when deployed on an expressly heterogeneous and noisy "virtual pooled" testing set with images originating from 50 different institutions \cite{neuroblastoma_DNE}.
    \item Ability to perform intricate and nuanced tasks, such as predicting change in overall tumor burden over time as class categories \cite{stember2022direct}. 
\end{itemize}

DNE's order of operations enables the above qualities because the CNN weights are able to sample broad swaths of the parameter space. In particular, DNE proceeds in the following basic sequence, repeated over and over again throughout training:
\begin{enumerate}
    \item \textbf{First}: \textit{update} the CNN weights randomly.
    \item \textbf{Then}: \textit{evaluate} the CNN by way of a selection criterion, here simply the total accuracy, summed over all training set images. 
\end{enumerate}
Due to the update-then-evaluate order, DNE's weight updates are not influenced by prior weights or model performance. Being thus unencumbered provides greater freedom to explore the parameter space of possible weights. Enhanced weight sampling supplies more flexibility and a greater range of possible functions that the CNN can feasibly approximate. The weight exploration can avoid the pitfalls of local loss surface minima and saddle points. 

In contrast to DNE, SGD works in reverse order, \textbf{first} \textit{evaluating} the loss after a forward pass, and \textbf{then} \textit{updating} the network weights via negative gradient. The resulting constraints on possible combinations of weight values often leaves SGD trapped in local loss minima or high-dimensional saddle points, making it nearly impossible to find or converge upon a global minimum.

\begin{figure}[H]
\includegraphics[width=12cm]{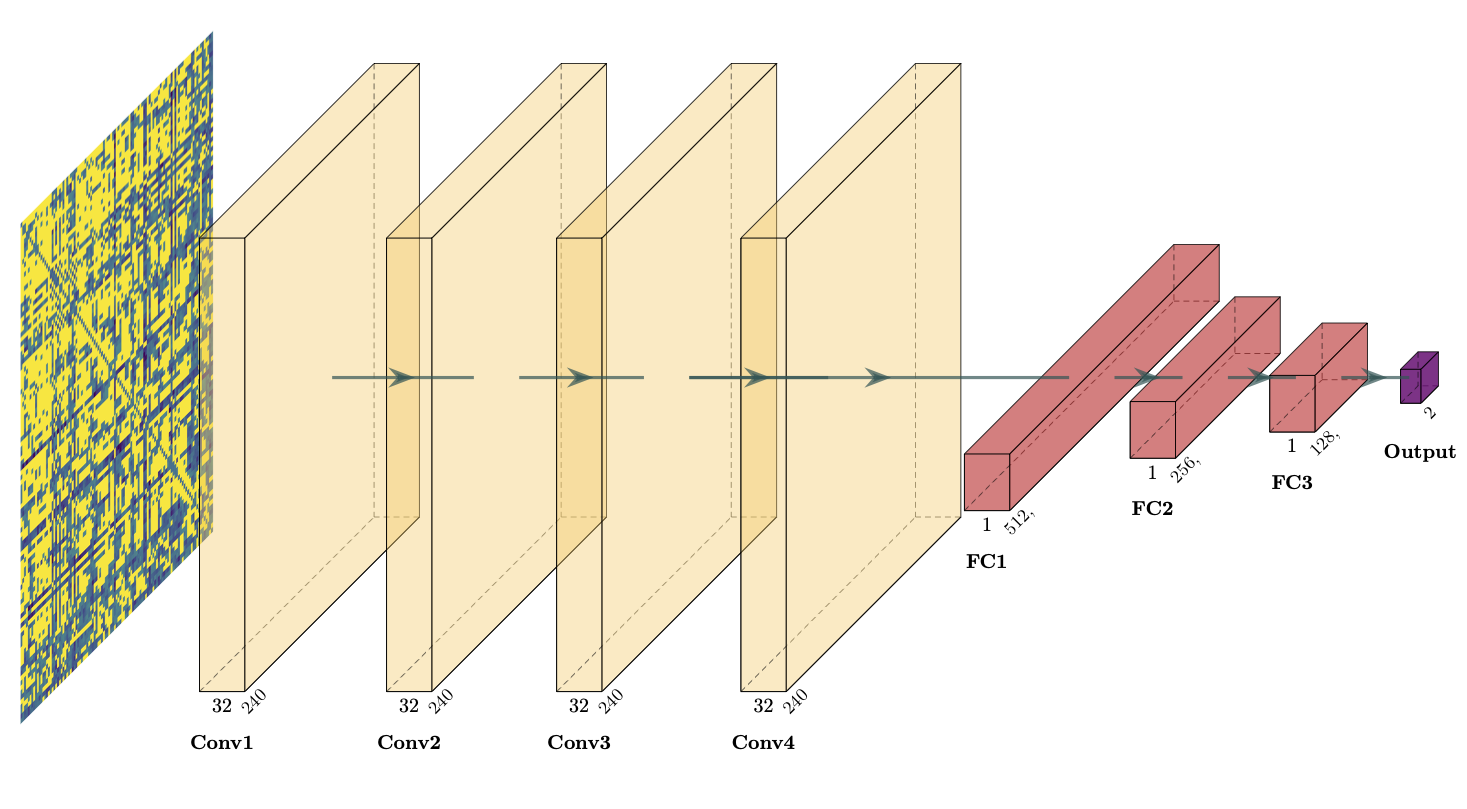}
\caption{Schematic of the convolutional neural network (CNN) architecture, with sample input adjacency matrix.}
\label{fig:CNN_architecture}
\end{figure}

The CNN that we used for this study, shown in Figure \ref{fig:CNN_architecture}, is fairly commonplace in overall structure and size. The network consists of 4 convolutional-relu layers, with 32-channel outputs, followed by flattening of the output maps and fully connected layers of size 512, then 256 then 128 nodes, all followed by selu activation. The output layer consists of 2-nodes, passed through a softmax function to normalize the sum of probabilities to one, hence producing an output.  Written more formally, given an input adjacency matrix $A$ and CNN represented by the function $F$ with trained weights $\vec{w}$, the output $O$ is obtained by 
\begin{equation}
O = F_{\vec{w}}(A).
\end{equation}
The predicted class corresponds to the node of highest probability, i.e. the argmax of the output. We employ labels of zero for LGG, and one for HGG. Hence, in the Python numbering system, the predicted class is 
\begin{equation}
    \text{predicted class} = 
    \begin{cases}
        \text{LGG} & \text{if argmax}(O) = 0 \\
        \text{HGG} & \text{if argmax}(O) = 1 \text{.}
    \end{cases}
\end{equation}
We employed $3 \times 3$ kernels, with weights randomly initialized according to the standard Glorot/Xavier scheme. Stride was 2 in both directions, with zero padding of one. Instead of anatomic images, the CNN inputs were simply the adjacency matrices. All calculations were performed with the built-in CPU for Google Colab Pro+. 

\section{Results}

\begin{figure}[H]
\includegraphics[width=12cm]{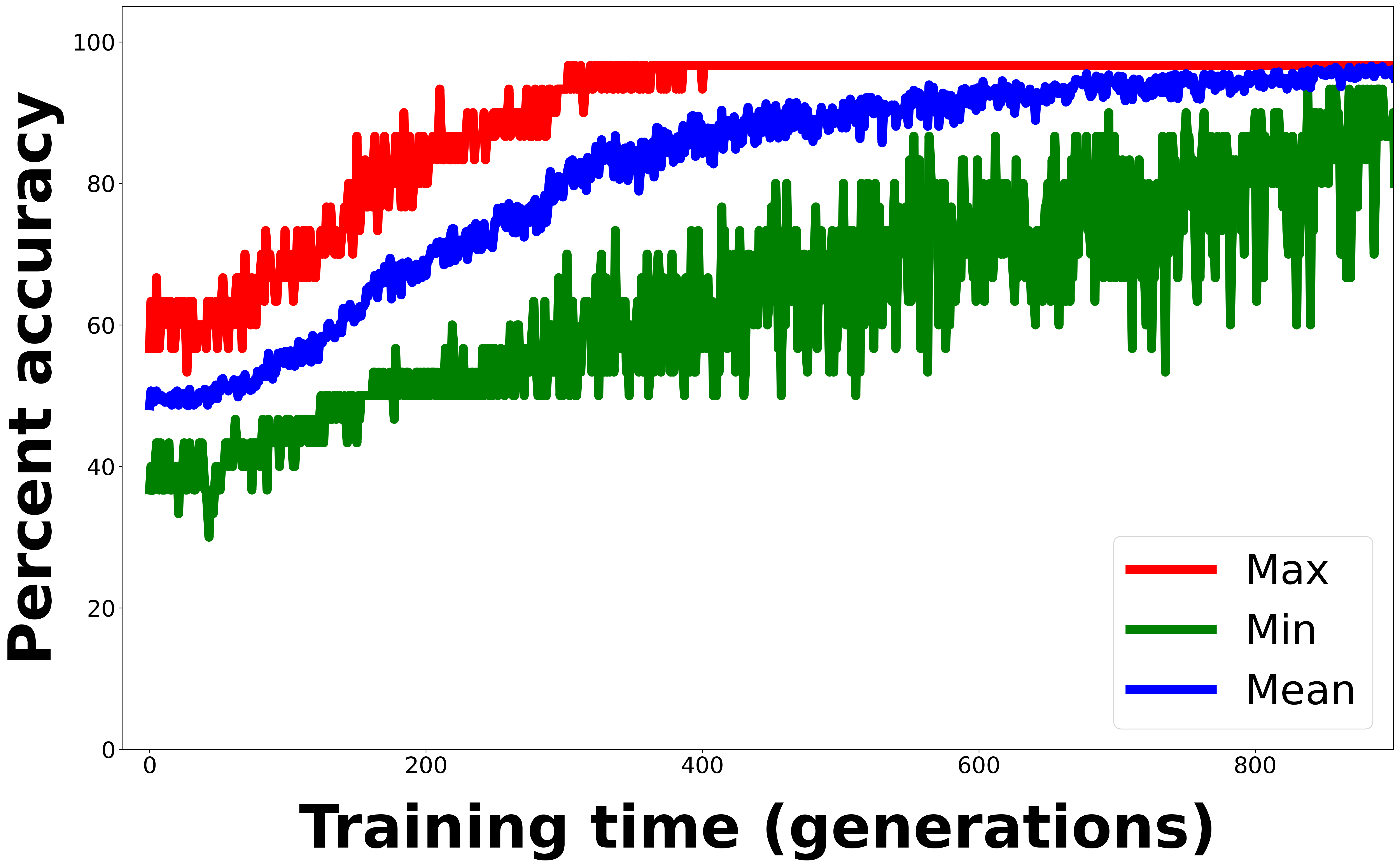}
\caption{Training set accuracy as a function of time in units of generations. The accuracies are for the child CNNs. The best-performing child (red) is plotted along with the average of children (blue), as well as the worst-performing child (green.) All three accuracies ultimately converge on high values.}
\label{fig:training_acc}
\end{figure}

Training time was set for an upper limit of 10,000 generations, but the runtime expired after around 2,000 generations. Nevertheless, convergence was achieved within 500 generations, as shown in Figure \ref{fig:training_acc}. Of note, one generation is equivalent to a full cycling of new child CNN weights, where they are updated in a positive and negative random direction each episode, evaluated once per episode, so that we have 40 episodes per generation. The parent CNN incorporates the best 50 percent of child network weights in each episode. 

\begin{figure}[H]
\includegraphics[width=12cm]{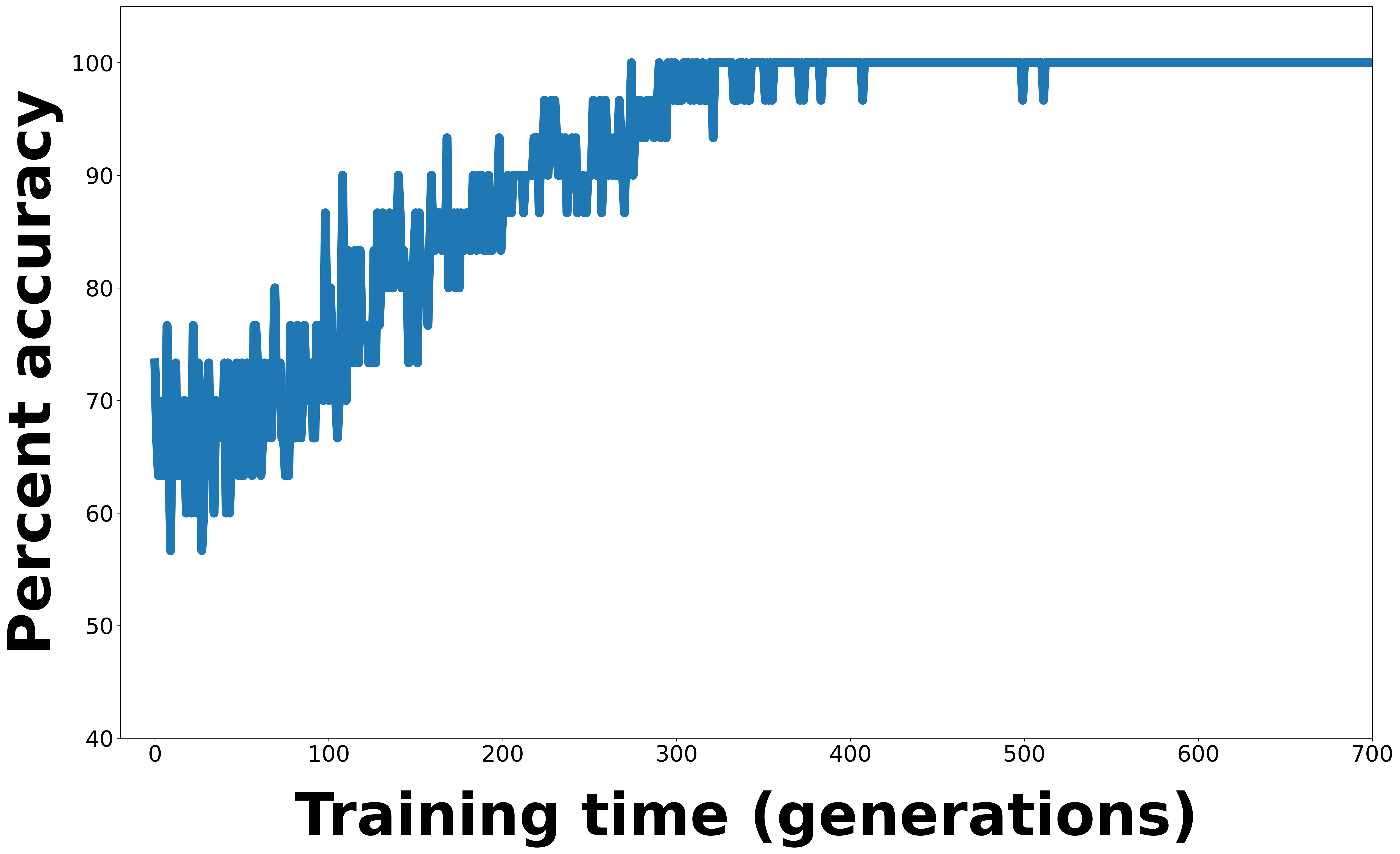}
\caption{Testing set accuracy as a function of training time.}
\label{fig:testing_acc}
\end{figure}

Of note, initial training was performed with a standard deviation for weight perturbation of 0.05, which had converged in other DNE work \cite{stember2021_DNE_sequence,stember2022direct} for anatomic images from the same data distribution (all from BraTS database or from the same institution). However, here the data is noisier and more heterogeneous, and thus training and testing set accuracy diverged for perturbation standard deviation of 0.05. Thus, we increased the perturbation standard deviation to 0.1. Doing so produced a broader range of weight perturbations. Since the vast majority of higher magnitude perturbations are not statistically expected to improve accuracy, training time increases. In sampling more weights, though, we improve generalization. In fact, accompanying the training set convergence to 100\% accuracy (Figure \ref{fig:training_acc}) was robust generalization to the testing set, which also arrived as 100\% accuracy, as shown in Figure \ref{fig:testing_acc}. 

\begin{figure}[H]
\includegraphics[width=12cm]{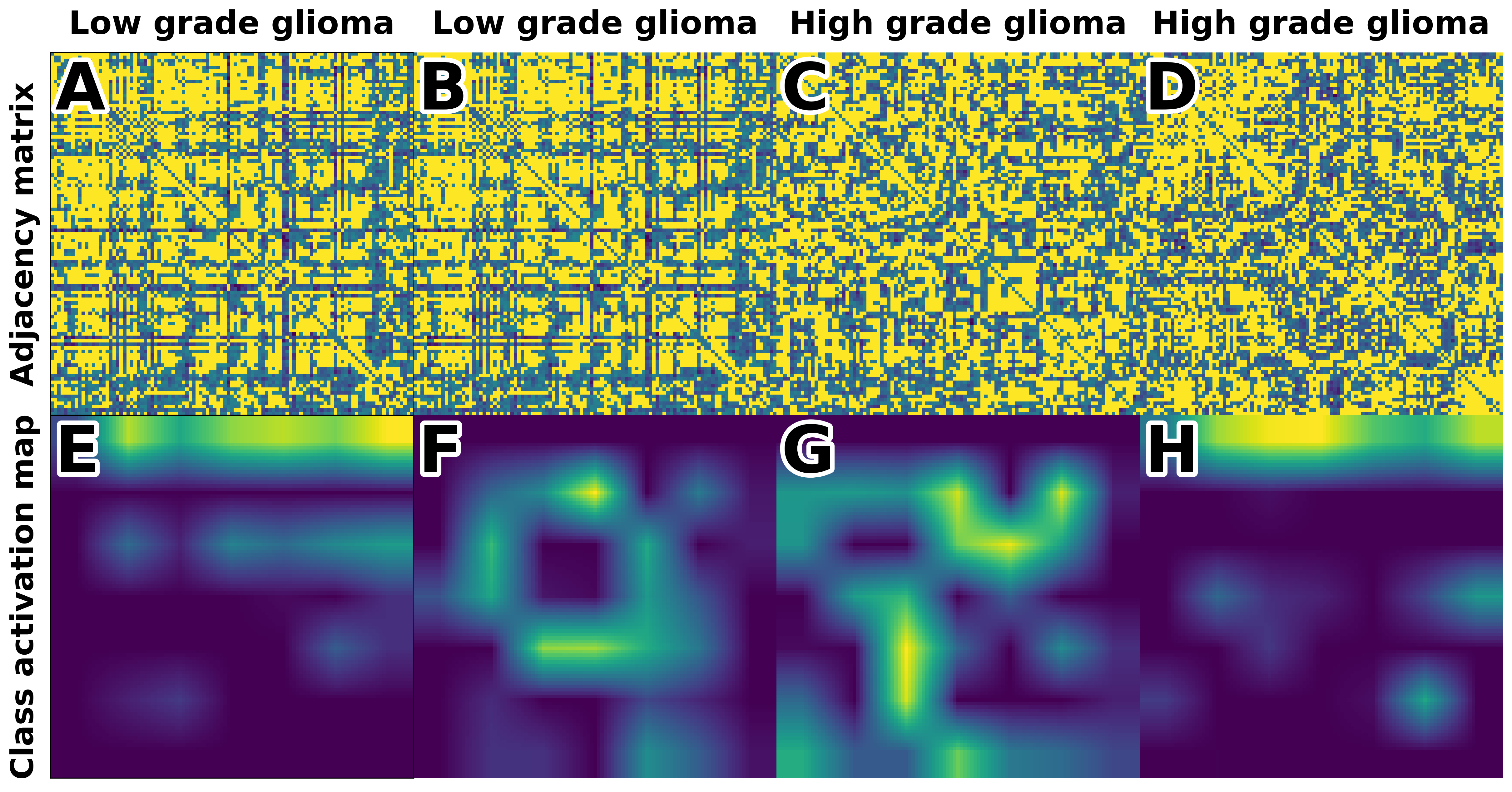}
\caption{Examples of adjacency matrices and their correspond CAMs}
\label{fig:CAM_egs}
\end{figure}

Figure \ref{fig:CAM_egs} shows a few testing set saliency maps, called class activation maps (CAMs). Whereas the CAMs for anatomic imaging may show a "hot spot" of focus on a lesion that determines malignancy, here the CNNs had to learn more sophisticated features given the data and task complexity. As a result, the CAMs have two major components, one that we may interpret, and the other that exceeds our intuition:
\begin{enumerate}
    \item Interpretable -- A dominant motif in the CAMs is one or more lines of focus along either rows or columns. This makes intuitive sense because it means that the network is looking at particular nodes and examining functional connectivity strength to other nodes.
    \item Less interpretable -- We see higher-order structures of saliency, notably for example in Figure \ref{fig:CAM_egs}F and \ref{fig:CAM_egs}G. Here a type of linear combination of the focus along connections of given nodes is present, along rows and/or columns, also with a spreading factor or variance present. 
\end{enumerate}

\section{Discussion}

We have seen that DNE can accurately predicting high- versus low-grade glioma based on resting-state functional MRI-derived adjacency matrices. DNE has in earlier work been shown to learn effectively on small and/or noisy anatomic images, and achieve the complex tasks of predicting change between successive scans. Here, we extend its repertoire to include arcane data that is not readily interpreted by humans, particularly if one had to predict tumor type based on the matrices. As it rises to meet this task, we see that DNE has to learn complex features, themselves only partially penetrable by us. 

Because we have shown a proof-of-principle ability of DNE to classify complex data forms, the approach could be applied to study other complicated forms of information, such as EEG or EKG recordings. In radiology, we could analyze the raw and unprocessed data forms of sinograms and k-space that undergo mathematical transformation into the human-interpretable CT and MRI scans, respectively. Being able to study the raw data could be helpful by avoiding the information loss that accompanies production of anatomic images. 

\subsection{Future directions} 

We would like to generalize the approach such that we can predict three or more classes, an obvious first step being the inclusion of normal healthy patients' adjacency matrices. Another important disease state to consider for cancer imaging is that of brain metastases.

While interesting to see that our CNNs learned complex features, as manifested by their ultimately inscrutable CAM maps, we seek more intuition whenever possible. To this end, future work will use fMRI data characterizing regional characteristics, such as BOLD asynchrony. A benefit of the regional type of fMRI is more intuitive saliency foci that correlate spatially with key anatomic structures.

\section{Conflicts of interest}

The authors have pursued a provisional patent based largely on the work described here.

\section{Funding}

We gratefully acknowledge external support from the Radiological Society of North America (RSNA) and the American Society of Neuroradiology (ASNR) as well as internal institutional research funding.

\section{Figures}

\printbibliography

\end{document}